\pdfoutput=1

\documentclass[11pt]{article}

\usepackage[]{latex/acl}

\usepackage{times}
\usepackage{latexsym}

\usepackage[T1]{fontenc}

\usepackage[utf8]{inputenc}

\usepackage{microtype}

\usepackage{inconsolata}

\usepackage{inconsolata}
\usepackage{pifont}
\usepackage{algorithm}
\usepackage{algorithmic}
\usepackage[namelimits]{amsmath} 
\usepackage{amssymb}             
\usepackage{amsfonts}            
\usepackage{mathrsfs}            
\usepackage{graphicx} 
\usepackage{subfigure} 
\usepackage{xcolor}
\usepackage{makecell}
\usepackage{booktabs}
\usepackage{multirow}
\usepackage{inconsolata}
\usepackage{pifont}
\usepackage{algorithm}
\usepackage{algorithmic}
\usepackage[namelimits]{amsmath} 
\usepackage{amssymb}             
\usepackage{amsfonts}            
\usepackage{mathrsfs}            
\usepackage{graphicx} 
\usepackage{subfigure} 
\usepackage{xcolor}

\usepackage{makecell}
\usepackage{booktabs}
\usepackage{multirow}
\usepackage{enumitem} 

\usepackage{newfloat}
\usepackage{listings}

\usepackage{xspace}
\newcommand{\SUM}{DIAC-Sum\xspace}
\newcommand{\QA}{DIAC-FactQA\xspace}

\definecolor{green}{RGB}{36, 214, 36}
\definecolor{red}{RGB}{235, 30, 30}

\usepackage{CJK}
\usepackage{xcolor}
\usepackage{ulem}

\definecolor{MyYellow}{rgb}{254, 246, 170}
\definecolor{MyBlue}{rgb}{170, 217, 251}

\title{Exploring the Factual Consistency in Dialogue Comprehension of Large Language Models}

\author{
    Shuaijie She \text{,} \textbf{Shujian Huang}\thanks{\quad Corresponding author} \text{,} \textbf{Xingyun Wang} \text{,} \textbf{Yanke Zhou}\text{,} \textbf{Jiajun Chen} \\
     \text{National Key Laboratory for Novel Software Technology, Nanjing University}  \\
    \normalsize\texttt{shesj@smail.nju.edu.cn}, \normalsize\texttt{huangsj@nju.edu.cn}\\
    \normalsize\texttt{\{wangxingyun,zhouyk\}@smail.nju.edu.cn},
    \normalsize\texttt{chenjj@nju.edu.cn} \\
}
\begin{document}
\maketitle
\begin{abstract}
LLMs (Large Language Models) usually interact with users in the form of dialogue and generate responses following their instructions, which naturally require dialogue comprehension abilities. However, dialogue comprehension is a general language ability which is hard to be evaluated directly.
In this work, we propose to perform the evaluation focusing on the factual consistency issue with the help of the dialogue summarization task.
Besides evaluating and analyzing the dialogue summarization performance (\SUM) of different LLMs, we also derive factual questions from the generated summaries and use them as a more flexible measurement of dialogue comprehension (\QA).
Our evaluation shows that, on average, 26.8\% of the summaries generated by LLMs contain factual inconsistency. Even ChatGPT, the strongest model evaluated, has such errors in 16\% of its summaries. For answering the factual questions, which is more challenging, the average error rate of all evaluated LLMs is 36.1\%. Both results indicate serious deficiencies. Detailed analysis shows that the understanding of subject/object of the conversation is still challenging for LLMs. Furthermore, to stimulate and enhance the dialogue comprehension ability of LLMs, we propose a fine-tuning paradigm with auto-constructed multi-task data, which achieved a relative error rate reduction of 11\% on \QA.\footnote{ We will release all data public to facilitate future research}
\end{abstract}

\section{Introduction}

With the development of large language models (LLMs), such as GPT-3~\cite{gpt3}, OPT~\cite{zhang2022opt}, LLaMA~\cite{touvron2023llama}, etc., it has becoming a promising way to interact with users through conversations, where LLMs generate responses following users' instructions in the dialogue. 
This form of conversational communication naturally requires high dialogue comprehension ability to capture the factual information, which has becoming a prerequisite for successfully completing tasks in the conversation.  

Previous studies have evaluated the dialogue comprehension ability of relatively small language models using dialogue question answering (QA) data, e.g., DREAM\cite{dream} and FriendsQA\cite{Yang2019FriendsQAOQ}. However, the questions in these data come from either human exams or random sampling of the dialogue content, neither of which targets on the \textbf{DIA}logue \textbf{C}omprehension (\textbf{DIAC}) ability of LLMs.

To fill this gap,  we propose to perform the evaluation with the help of the dialogue summarization task (\SUM), since summarization extracts important information from the dialogue which naturally requires a correct understanding of the dialogue. The summaries generated by 5 popular LLMs are collected and manually evaluated for factual consistency. The results are reported as an empirical study of the current situation. 

More importantly, we derive a set of factual questions from the factual inconsistencies in the generated summaries (\QA), which corresponds to the difficult parts of understanding these dialogues. By answering these questions, the dialogue comprehension ability of LLMs could be evaluated in a more flexible and precise way. Please note that the resulting dataset could be used for evaluating the dialogue comprehension ability of other LLMs in the future, without any more human annotation.

\begin{table*}[htbp]
\setlength{\tabcolsep}{1mm}
\centering
\footnotesize
\begin{tabular}{l|r|p{10cm}}\toprule
\multirow{5}{*}{\textbf{Dialogue:} } & \multicolumn{2}{l}{Marsha: Guys, we've planned the trip with John last night as we promised.}  \\
  &\multicolumn{2}{l}{ Cynthia: great, thank you for that.}\\
  &\multicolumn{2}{l}{ Marsha: but of course you have to agree on that.}\\
& \multicolumn{2}{l}{Mohammad: sure, but I really trust you.}\\
 & \multicolumn{2}{l}{Gavin: me too.} \\
 & \multicolumn{2}{l}{\quad \dots} \\\midrule \midrule
 \multirow{5}{*}{\textbf{\SUM}}  & \textbf{Erroneous Summary:}  & Marsha, Cynthia, Mohammad and Gavin are going to Madagascar. \\\cmidrule{2-3}
 & \textbf{Error:} &  The participation of John is neglected.  \\
 & \textbf{Error Type:} & SubObjE  \\
\cmidrule{2-3}
 & \textbf{Corrected Summary:} & Marsha, Cynthia, Mohammad, Gavin and John are going to Madagascar.\\
 \midrule 
 \midrule
 \multirow{9}{*}{\textbf{\QA}} & \textbf{Factual Question 1:} & Who is going to Madagascar?\\
 &Correct Answer: & Marsha, Cynthia, Mohammad, Gavin and John\\
  & Wrong Answer: & Marsha, Cynthia, Mohammad and Gavin\\
\cmidrule{2-3}
& \textbf{Factual Question 2:} & How many people are going to Madagascar?\\
& Correct Answer: & 5\\
& Wrong Answer: & 4\\
\cmidrule{2-3}
 & \textbf{Factual Question 3:}&  Is John going to Madagascar?\\
 & Correct Answer: &  yes\\
& Wrong Answer:  & no\\
\bottomrule
\end{tabular}
\caption{\label{tab:annotation} This is an example of our annotation process. The Dialogue is a partial input dialogue from the SAMSUM dataset. The Error Summary is the output of fine-tuned the BART-Large model on this dataset which has SubObjE (missing Subject John). The Corrected Summary is the summary that the annotator has revised with minimal changes. Factual Question 1-3 are constructed by the annotator to test the dialogue comprehension capability. 
}
\end{table*}

Our evaluation results show that existing LLMs still have serious deficiencies in conversational understanding. 
On average, 26.8\% of the summaries generated by LLMs contain factual inconsistencies. Even the summaries generated by ChatGPT, the strongest LLMs in our evaluation, has 16\% identified to be incorrect. Further factual question probing results shows that the average error rate of all evaluated LLMs reaches 36.1\%. For the reference, the error rate of ChatGPT and Vicuna-13B is 26.2\% and 40.5\%. Detailed analysis on different error categories show that the understanding of subject-object is one of the most challenging problem.

We also present attempts to stimulate the dialogue comprehension ability by fine-tuning the model with auto-constructed multi-task data. 
Experimental results demonstrate that after fine-tuning, the model exhibits a better dialogue comprehension ability, providing a potential direction for future work and improvements.


\section{Dialogue Comprehension Benchmark (DIAC)}
\subsection{Data and Model Preparation}
Our evaluation utilized the SAMSum~\cite{samsum} dataset, which is composed of 
messenger-like conversations together with their summaries. These conversations cover various topics in real-life messenger chats, ranging from informal to formal and showcasing the use of slang words, emoticons, and typos. This format bears resemblance to the current interaction between users and LLMs. The summaries succinctly summarize the overall conversation in third person. We use the same subset consisting of 150 conversations from the test set of SAMSum as~\citealp{data}, so our analysis could also be compared with their results.

Our evaluation is based on the generations of the LLMs, and five popular LLMs are selected to initialize the evaluation: Alpaca-7B\footnote{https://github.com/tatsu-lab/stanford\_alpaca}, Vicuna-7/13B\footnote{https://github.com/lm-sys/FastChat\#vicuna-weights}, Flan T5-11B\footnote{https://huggingface.co/google/flan-t5-xxl}, and ChatGPT\footnote{https://platform.openai.com/docs/models/gpt-3-5}. These models vary in size, training methods, and training datasets, but all of them have achieved impressive results and are widely used currently. 
Latest LLMs such as Mistral~\cite{jiang2023mistral}, LLaMA2~\cite{touvron2023llama2} and GPT4 can also be incorporated into our evaluation in an annotation-free style.



\subsection{\SUM: Inconsistency Annotation and Correction}

As summarization requires a comprehensive understanding of the information presented in the conversation,  inconsistencies in the summary may indicate incorrect comprehension of the dialogue. 

We ask the initial five LLMs to generate summaries for each conversation, resulting 750 summaries. For the purpose of evaluating factual consistency, we require the LLMs to generate summaries that contains exact information in the conversation rather than some vague description. This is achieved by a sampling and filtering strategy. Please refer to the Appendix~\ref{sec: filter rule} for details.

We manually annotate and correct the factual inconsistencies in the collected summaries according to \citealp{data}, where the inconsistencies are categorized into the following five types. 

\begin{itemize}[topsep=2pt, itemsep=3pt, parsep=0pt, partopsep=0pt]
    \item \textbf{SubObjE}: additions, deletions, or substitutions of participants (as subjects or objects) described in the summary. 
    \item \textbf{ProE}: incorrect references of pronouns in the summary. 
    \item \textbf{HalE}: events in the summary not exist in the dialogue, resembling a hallucination.
    \item \textbf{ParE}: particular errors in the time and location of events. 
    \item \textbf{NegE}: contradictions in the summary with events in the original dialogue. 
\end{itemize}


In the manual annotation process, each annotator is assigned with a dialogue and a corresponding summary generated by LLMs. The annotation task is to verify the consistency between the generated summary and the dialogue, and to identify the error along with its type, if present. The annotator is also asked to correct the error summary with minimum editing efforts.
The annotation process involves 5 annotators with an agreement of coefficient $\textit{k} = 0.83$. All 750 summaries are annotated. The performance of each LLMs are evaluated by the factual errors in their respective generated summaries. More details can refer to Appendix~\ref{sec: Annotation Details}.

An example of the annotation of \SUM is shown in Table~\ref{tab:annotation}. Given the conversation, the error summary neglects John's participation in the trip. According to the definition, this inconsistency is classified as SubObjE and the participation of John is integrated into the summary for correction.

\begin{table}[t]
\setlength{\tabcolsep}{2mm}
\footnotesize
\centering
\begin{tabular}{lccccccl}\toprule
\textbf{Settings} &  \textbf{\QA}  & \textbf{DREAM}\\\midrule
\text{Dialogue Length (tokens)} & 106.0 & 53.8 \\
\text{Dialogue Turns} & 	12.7	& 4.4  \\
\text{Speaker Num} & 2.5 & 2.0 \\
\text{Question Length (tokens)} & 8.5&	6.9 \\
\text{Questions per dialogue} & 10.0 & 1.2 \\
\text{Dialogue Distinct 1-gram} & 0.17 & 0.08\\
\text{Dialogue Distinct 2-gram} & 0.67	& 0.42\\
\bottomrule
\end{tabular}
\caption{\label{DatasetDetails} We compile some basic characteristics of the dataset and compare them with DREAM. Distinct~\cite{distinct} is a common metric used to measure the diversity of a text by the repetition of n-grams; the higher it is, the better the diversity of the text.}
\end{table}

\subsection{\QA: Factual QA Construction}

The result of previous annotation presents the ability of current LLMs, however, the annotation process is expensive and cannot be reused directly.
Therefore, we present \QA, which transfer the evaluation of summaries into a question-answering process. Different from previous practice that also using QA to evaluate the dialogue comprehension abilities~\cite{dream,Yang2019FriendsQAOQ}, our \QA focuses on inconsistencies in the auto-generated summaries, so it is more accurate in identifying the weaknesses in the model's understanding of the dialogue.



More specifically, we ask the annotators to compare the corrected summary with the error summary and wrote several questions based on the previously identified errors. For each question, two choices of answers are collected, including answers from both the correct summary and the error summary. This process yields a correct answer and a distractor associated with each question, with the dialogue serving as a reference. 

In order to ensure comprehensive testing of the model's dialogue comprehension, the annotators are required to create multiple types of questions for each error, as shown in Table~\ref{tab:annotation}. In this example, the model overlooks the fact that John is also participating in the trip. To test whether the model actually knows the fact, we annotate three diverse questions from different perspectives, including a question for listing all the participants, a question for the number of participants and a direct question for John's participation. 

Besides \SUM we construct, ~\citealp{data} also collected 750 summaries, including the result generated by four BART-based models and the original reference summary in SAMSum. For better diversity, we also corrected these summaries and merged two datasets to obtain 1,500 summaries for our \QA annotation, resulting in a total of 446 triples of (dialogue, erroneous summary, corrected summary). 
In total, we collected 1484 questions together with two options, answer and distractor, based on the inconsistent summaries.  On average, an inconsistent summary has 3.32 questions.

\begin{table*}[t]
\setlength{\tabcolsep}{6mm}
\footnotesize
\centering
\begin{tabular}{l|ccccc|c}\toprule
\textbf{Model}  & \textbf{SubObjE} & \textbf{ProE} & \textbf{HalE} & \textbf{ParE} & \textbf{NegE}  & \textbf{Overall} \\\midrule 

Alpaca-7B    & 16.0 & 4.0  & 18.7  &  12.7  & 1.3  & 40.0 \\
Vicuna-7B   & 14.0   & 7.3 & 12.0 &  10.7  & 5.3   & 34.7\\
FlanT5-11B  & 9.3  & 5.3 & 2.7  &  2.0 & 4.0  & 18.0  \\
Vicuna-13B  & 10.7  & 5.3  & 7.3 &  4.0  & 4.0  & 25.3 \\
ChatGPT     & 4.7       & 5.3 & 2.7  &  4.7  & 0.7   & 16.0\\
GPT4     & 4.0	& 3.3&	3.3	& 1.3&	0.7	& 11.3\\ \midrule
Average & 9.8 & 5.1 & 7.8 & 5.9 & 2.7 & 24.2 \\ \midrule
Human-Ref*   & 2.7     & 3.3  & 0.7 &  2.0  & 1.3 & 9.3\\
BART-based*   & 14.7    & 6.0  & 5.3 &  14.7  & 6.0  & 36.7 \\
\bottomrule
\end{tabular}
\caption{\label{SUMFact} The proportion of inconsistencie ($\%$) of \SUM. The row of Average is the average of all 5 LLMs. The rows of Human-Ref and BART-based models come from \citealp{data}, for comparison purposes.}
\end{table*}

\begin{table*}[t]
\setlength{\tabcolsep}{4mm}
\footnotesize
\centering
\begin{tabular}{l|ccccc|c||c}\toprule
\textbf{Model} & \textbf{SubObjE} & \textbf{ProE} & \textbf{HalE} & \textbf{ParE} & \textbf{NegE}  & \textbf{Overall} & \textbf{DREAM} \\\midrule 
Alpaca-7B   & 51.6 & 47.5 & 58.0 & 47.7 & 46.6  & 48.5 & 31.0\\
Vicuna-7B   & 47.0 & 46.5 & 49.0 & 46.4 & 48.3 & 46.3 & 32.5 \\
FlanT5-11B   & 25.5 & 18.8 & 26.4 & 30.2 & 27.5 & 24.8 & 4.80 \\
Vicuna-13B  & 41.6 & 39.0 & 42.0 & 43.4 & 32.8  & 40.5 & 17.9 \\
ChatGPT     & 25.2 & 24.6 & 29.8 & 28.1 & 31.4 & 26.1 & 6.10 \\ \midrule
LLaMA2-7B-Chat  & 48.1 & 48.0 & 51.5 & 50.6 & 45.1 & 47.7 & 29.1 \\
Mistral-7B-Instruct  & 38.7 & 37.0 & 42.0 & 36.2 & 31.7  & 36.7 & 14.5 \\
GPT4     & 15.2 & 14.0 & 24.2 & 20.1 & 22.2 & 18.5 & 3.32 \\\midrule
Average   & 36.6 & 34.4 & 40.4 & 37.8 & 35.7  & 36.1  & 17.4 \\

\bottomrule
\end{tabular}
\caption{\label{QARESULT} Error rate $(\%)$ of answering factual questions on \QA and DREAM.}
\end{table*}

Table~\ref{DatasetDetails} shows some basic information about our dataset. DIAC-FactQA has higher distinct n-gram, offering better diversity compared to DREAM. Moreover, DIAC-FactQA includes longer dialogues and more turns per conversation, increasing its complexity. With a larger set of factual questions, DIAC-FactQA allows for a more comprehensive assessment of dialogue understanding consistency. More details about our dataset can refer to Appendix~\ref{Distribution}.

\section{Evaluation Results and Analysis}
\subsection{\SUM Results and Analysis}

Table~\ref{SUMFact} shows the factual consistency of LLMs on the summarization task. ``BRAT-based'' are the results of BART-based models after supervised training  for summarization; ``Human-Ref'' refers to the summary written by human. These two results are adopted from \citealp{data}. The main observations are as follows.

\noindent \textbf{Inconsistency still plague LLMs:}
ChatGPT has the best performance, with the overall error rate and each individual error categories being better than those of other models. However, surprisingly, 16\% of the summaries still contain inconsistency. This indicates that even on the top-performing ChatGPT, there are still many errors in generation. It is worth noting that FlanT5 was trained on the SAMSum dataset and thus has better performance. In addition, Alpaca and Vicuna were both trained on the instruction datasets without seen this summarization data directly, with error rates ranging from 25.3\% to 40\%. The average inconsistency rate in the summaries generated by all the models is 26.8\%. The problem of factual inconsistency remains a serious issue, which also points to the existing weakness in the comprehension of dialogues by LLMs.

\noindent \textbf{Stronger language abilities reduce errors:}
Compared with the previous BART models, the LLMs make fewer errors. For example, FlanT5, which is also trained on SAMSum, makes far fewer errors than the BART model, from 36.7\% to 16.8\%. 
We also observed results that are similar to other evaluation benchmark~\cite{zheng2023judging,open-llm-leaderboard}, where Vicuna 7B showed better performance than Alpaca 7B. Moreover, with parameter scaling up, Vicuna 13B demonstrated even better results. The model's better performance and dialogue comprehension ability are related.

\begin{table*}[ht]
\setlength{\tabcolsep}{2mm}
\footnotesize
\centering
\begin{tabular}{l|l}\toprule
\textbf{Task Re-Assignment} & (SubObjE) \\\midrule
Maya: Bring home the clothes that are hanging outside & Who will bring home the clothes? \\Boris:  I'm not home right now & ChatGPT: B: Boris \\
Boris: I'll tell Brian to take care of that & Correct: A: Brian \\\midrule\midrule
\textbf{Intention of Participation} & (SubObjE)\\\midrule
Marsha: Guys, we’ve planned the trip with John last & How many people are going to Madagascar? \\night 
as we promised & ChatGPT: B: Four \\
Cynthia: great, thank you for that & Correct: A: Five  \\
Marsha: but of course you have to agree on that &  \\
Mohammad: sure, but I really trust you & \\
Gavin: me to &  \\\midrule \midrule
\textbf{Complex Pronoun References} & (SubObjE, ProE)\\\midrule
Ian: I don’t know any Claire... &  Who did Ian dump years ago?\\ 
Maddie: Really? I spoke with Leah and she told & ChatGPT: B: Leah\\me that
you dumped her years ago & Correct: A: Claire \\
Ian: are you sure she was talking about me? believe me \\
none was named Claire   & \\\midrule \midrule
\textbf{Subsequent Clarification} & (SubObjE, HalE) \\\midrule
Timmy: What about food? & Who will cover the food at the BBQ? \\
Gemma: Others and I will cover it & ChatGPT: B: Gemma \\
Timmy: Others? I thought it was a date :P & Correct: A: Gemma and her boyfriend \\
Gemma: U remember I have a bf, right?  \\\midrule \midrule
\textbf{Inference Intent from Expressions} & (SubObjE, NegE and HalE)\\\midrule
Brian: lets NOT do the homework and see what happens & Is Lena not going to do the homework? \\
Lena: do what you want I won’t risk it & ChatGPT: A: Yes \\
& Correct: B: No\\\bottomrule
\end{tabular}
\caption{\label{case study} Analysis of problematic dialogue comprehension cases from ChatGPT.
}
\end{table*}

\noindent  \textbf{SubObjE and HalE remain the major issues:}
Similar to BART, SubObjE remains challenging for LLMs. The SubObjE involves understanding the referring  information in the dialogue, analyzing the speaker's intention (such as whether to attend an event), handling abbreviations in the dialogue, and other issues. Another noteworthy observation is that the hallucination in the LLMs has increased significantly. How to make the large-scale model aware of the factual consistency and generate reliable output based on the input dialogue is a problem that requires further attention.

\subsection{\QA Result and Analysis}

The results of different LLMs on \QA are shown in Table~\ref{QARESULT}. Besides the initial 5 LLMs used for generating summaries, we also extend our evaluation to later LLMs such as LLaMA2, Mistral and GPT4. This extension does not require any further annotation.

\noindent  \textbf{Many LLMs have severe deficiencies in dialog understanding:} 
The experiment results on \QA confirm 
the deficiencies in dialogue understanding of LLMs. Even ChatGPT and GPT4 have error rates of 26.1\% and 18.5\%, respectively. In addition, Vicuna-13B, which is considered a relatively strong LLMs, has an error rate of 40.5\% on these questions. 
The problem still remains severe on latest models such as LLaMA2, Mistral, and GPT4. Especially for GPT4, which has extraordinary performance currently, there are still 18.5\% of questions that cannot be answered correctly. The average error of LLMs even reaches 36.1\%, revealing their limited understanding of conversations.

\noindent  \textbf{\QA has diagnosed more issues:}
Compared with DREAM, \QA can more effectively discover these conversation understanding defects. Large models can answer many test questions in the DREAM dataset. For example, Mistral, ChatGPT, and GPT4 has a low error rate of 14.5\%, 6.1\%, and 3.3\%. However, in \QA, their performance are much worse: 36.7\%, 26.1\%, and 18.5\% respectively. \QA contributes to a more comprehensive diagnosis of deficiencies in dialogue understanding and facilitates future improvements.

\begin{figure*}[ht]
\centering
\includegraphics[scale=0.47]{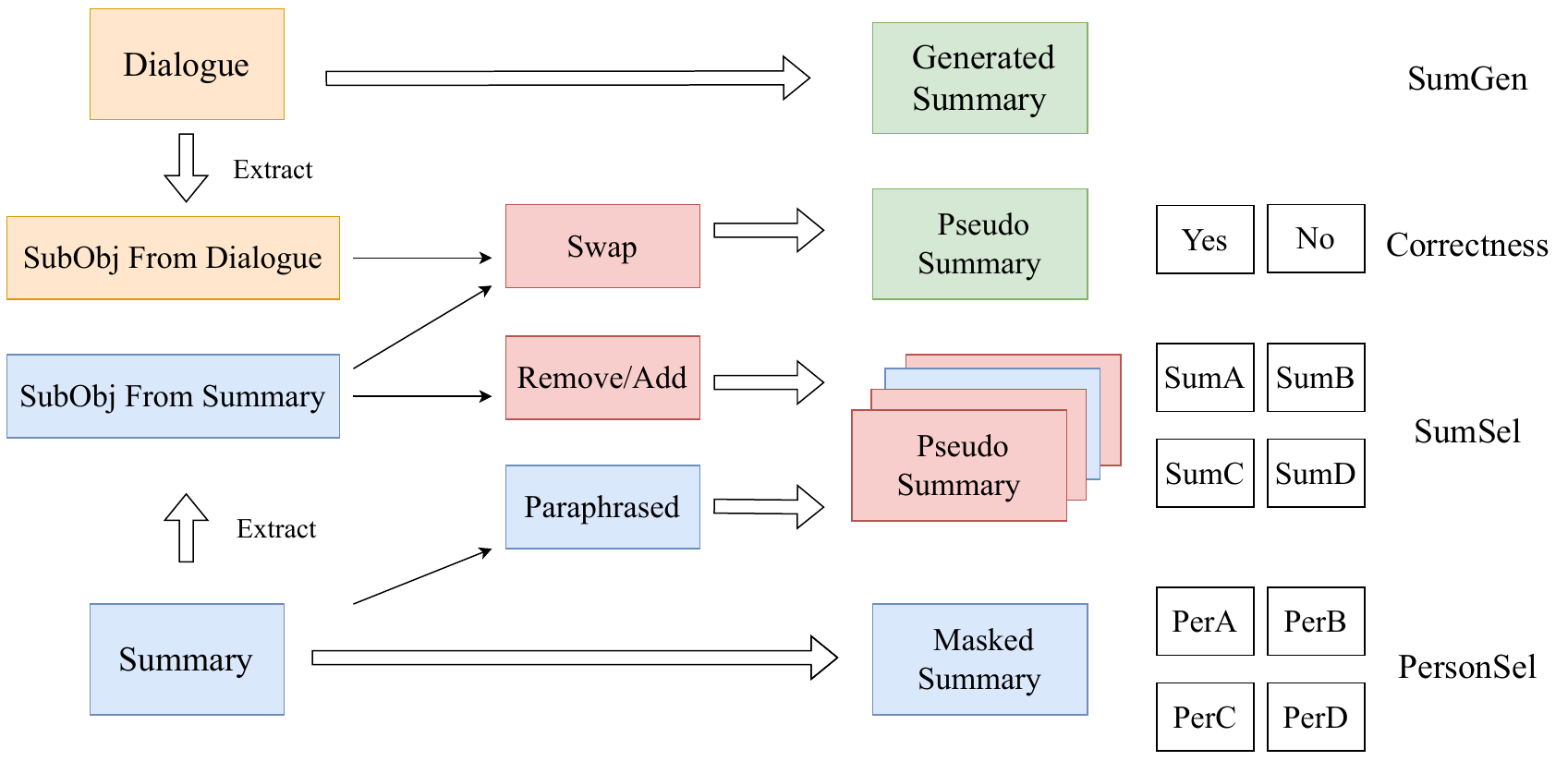}
\caption{Illustration of our multi-task pseudo-data. The negative examples, highlighted in red, represent constructed instances that contain subject-object errors. Conversely, the positive examples are indicated in blue.}
\label{pve}
\end{figure*}


\subsection{Case Study}

Previously, we analyze the dialogue comprehension ability of LLMs across inconsistent categories.
We also conducted an examination of specific scenarios in which models are susceptible to errors (Table~\ref{case study}). 

In daily conversations, there are often task assignments, activity arrangements, etc. When the individuals indicate that they are unavailable and have made \textbf{task re-assignments}, LLMs may not fully comprehend the information and neglect the most recent updates. This may result in ChatGPT mistakenly presuming that Boris will complete the task of gathering clothing. 

The \textbf{intention of participation} corresponds to activity organization commonly found in the dialogues. The model needs to integrate scattered information from the dialogue in order to correctly understand which people are participating in the activities. In this example, the model easily overlooks John and Cynthia. 

One notable characteristic of dialogues is the extensive use of pronouns to facilitate communication. The example involves two speakers and two individuals in the conversation, and include a significant number of \textbf{complex pronouns}. While humans can comprehend the dialogue, it poses a considerable challenge for the model. The answers of factual question show that the model's understanding of these pronouns are quite confusing.

Meanwhile, temporary uncertainties often arise during discussions, but these are typically resolved through \textbf{subsequent clarification} in the  dialogue. In the given example, it is explained later that ``Others" refers to the boy friend. However, it requires the casual language model to associate the later explanation with the earlier uncertainty, which is also a non-trivial challenge. 

\textbf{Inference intention} in dialogue presents another challenge. Speakers often use a variety of expressions to convey their intentions, such as the phrase ``I'm not one to take risks'' in the example.  It is crucial for the model to accurately interpret whether the user intends to express agreement or disagreement with the various stances presented.

\section{Multi-Task Fine Tuning Paradigm}
\subsection{Pseudo Dataset Creation}

The above analysis reveals that the primary deficiency of current LLMs in conversation understanding is the recognition of subject and object in dialogue. Therefore, we made some initial attempts to improve the dialogue understanding ability of LLMs by focusing on their subject and object comprehension capabilities.


Our intuition is simple: the model must correctly understand the subject-object in the dialogue to perform the tasks accurately. More important, in order to drive and validate the model's genuine learning of general dialogue understanding abilities, rather than simply fitting the distribution and task forms, the synthesized data we created differs from the evaluation data regarding its distribution, task structures, and instructions. Our data construction is based on the SAMSum dataset. Figure~\ref{pve} is an overview of our construction process. In addition, Table~\ref{example} provides an example of data construction. Further details, including the instructions we designed for each task, will be discussed in the 
Appendix~\ref{sec:instruction}.

\begin{table*}[t]
\setlength{\tabcolsep}{3mm}
\footnotesize
\centering
\begin{tabular}{c|p{12cm}}\toprule
\textbf{Input Dialogue} & Jenkin: hey what is your spirit animal? \\
& Sophie: what? \\
& Sophie: I dont know, a fox lol \\
& Jenkin: are you wiley? \\
& Sophie: sometimes \\
& Jenkin: I am a dolphin \\
& \dots \\\midrule 
\textbf{Candidate Summary}  & \dots Sophie would choose a fox. \\\midrule 
\textbf{SumGen} & 
 Jenkin have be read about spirit animals and he be draw to a dolphin. Sophie would choose a fox. \\
 \midrule
 \textbf{Correctness}  & The candidate summary is inconsistent or not? \textbf{Label: } A: No \\
 \midrule
 \textbf{SumSel} & Identify the correct summary of the dialogue from the provided options \\
& A: Jenkin have read about spirit animals and he be draw to a dolphin. Sophie would choose a fox. Jenkin and wiley will bring pack of cards with spirit animals to Sophie tomorrow.\\
& \dots \\
& D: Jenkin have read about spirit animals and he draw to a dolphin. Sophie would choose a fox. Jenkin will bring pack of cards with spirit animals to Sophie tomorrow.\\
& \textbf{Label: } D  \\\midrule 
\textbf{PersonSel} & Based on the conversation, select the most suitable option to fill in the [MASK] blank.\\
& \dots [MASK] would choose a fox.\\
&  A:  Jenkin \quad B: Sophie \quad C: Sophie and Jenkin \quad D: Jenkin\\
& \textbf{Label: } B: Sophie \\\bottomrule
\end{tabular}
\caption{\label{example} An example of multi-task pseudo data
}
\end{table*}

Summary is a brief overview of the core content in a conversation. To understand the dialogue, a clear comprehension of the subject and object in its summary is essential. Therefore, we propose the \textbf{PersonSel} (Person Selection) by masking the person in the reference summary, and requiring the model to select the correct option from four choices based on the dialogue to fill in the masked position.
Moreover, dialogue summarization is a comprehensive form of learning dialogue comprehension. We also sampled some dialogue summarization data for training as \textbf{SumGen} (Summary Generation).


Aside from training models to deduce correct answers, teaching models to recognize errors is also a viable approach. Models require the ability to comprehend dialogues and infer the correct subject-object in order to accurately identify errors. To obtaining negative examples, we extract subjects and objects from the dialogue and summary respectively, including speakers and person entities\footnote{https://spacy.io/models/en\#en\_core\_web\_lg}. For the summary's subject-object, we simulate SubObjE by substitution, addition, and deletion to create negative examples. Meanwhile, we can generate positive examples using a paraphrasing model.

Once we have obtained positive and negative examples, we can construct \textbf{Correctness} (Correctness Decision) data to teach model which is correct or not. We further simplified this task to \textbf{SumSel} (Summary Selection) by providing four options for the model to choose. One of these is correct, while the remaining three contain subject-object errors. As a re-ranking task, it enables the model to learn the required abilities in an easier way.

\subsection{Experimental Settings}

\begin{table}[t]
\setlength{\tabcolsep}{1.1mm}
\footnotesize
\centering
\begin{tabular}{lccccl}\toprule
 & \multicolumn{2}{c}{\textbf{\QA}} & \multicolumn{1}{c}{\textbf{DREAM}} \\\midrule 
\textbf{Model}   & \textbf{SubObjE}   & \textbf{Overall } & \textbf{Overall }\\\midrule 
Alpaca-7B   &  44.1(-7.95\%)  &  42.9(-11.5\%)  &  21.2(-31.6\%)  \\
Vicuna-7B    &  42.9(-8.72\%) &  42.0(-9.30\%)  &  22.6(-30.5\%)  \\
Vicuna-13B    &  34.4(-17.3\%)  & 35.6(-12.1\%) &  15.2(-15.1\%)  \\ \midrule
Average  &  40.5(-11.3\%)   & 40.2(-11.0\%)  &  19.6(-27.6\%)  \\ 
\bottomrule
\end{tabular}
\caption{\label{LoRATune Result1} Question answering error rate $(\%)$ on \QA and DREAM after multi-task fine-tuning. Error reduction rate is caculated relatively.
}
\end{table}

Based on the training set of the SAMSUM dataset, we automatically sampled 10,000 data for each task, with a total of 40,000. At the same time, we hope to enhance the ability of dialogue comprehension while maintaining the original ability of the model as much as possible. Therefore, we mixed pseudo-data and Alpaca's instruction-following data. For efficient fine-tuning, we used LoRA~\cite{hu2021lora} to fine-tune the model. Each model was trained for 3 epochs and the training process took only 1 hour on 8 A100 80G. We run three times of each model with different random seeds and report the average results.

\subsection{Fine Tuning Result and Analysis}
Table~\ref{LoRATune Result1} shows the performance of different LLMs after fine-tuning. 
Our data not only enhances the model's understanding of subject/object, but also improves the overall dialogue comprehension ability. It has achieved stable improvements on two datasets, which are 11\% and 27.6\% respectively.
As we designed, there exist notable difference between our training data and test data which highlights the efficacy of our data in enhancing the model's core ability of dialogue understanding, which can generalize across different tasks. 
Although LLMs have overcome more difficulties in dialogue understanding with the help of our data, it can be observed that there are still many errors. \QA remains a challenging tasks that requires further in-depth research in the future.

There is a reasonable gap between the performance in Table~\ref{LoRATune Result1} and FlanT5 in Table~\ref{SUMFact}.  We believe that this is mainly due to FlanT5 has been fully fine-tuned on a larger and more diverse training data set (including the SAMSum and DREAM and other QA data in a format similar to test set). 
\subsection{Ablation Study}


\begin{table}[t]
\setlength{\tabcolsep}{3mm}
\footnotesize
\centering
\begin{tabular}{lccccccl}\toprule
\textbf{Settings} &  \textbf{\QA}  & \textbf{DREAM}\\\midrule
\text{Baseline} & 48.5 & 31.0 \\
\text{All} & 42.9 & 21.2 \\\midrule
 \text{w/o SumGen} & 45.3 & 24.7 \\
\text{w/o Correctness } & 46.2 & 25.4 \\
\text{w/o SumSel} & 43.9 & 22.3 \\
\text{w/o PersonSel} & 46.0 & 28.5\\
\bottomrule
\end{tabular}
\caption{\label{Abaltion Study} Overall question answering error rate $(\%)$ on \QA and DREAM in ablation study.
}
\end{table}

In order to validate the impact of each task on the learning process, we remove one task at a time and fine-tune Alpaca 7B using the remaining data. The experimental result in Table~\ref{Abaltion Study} demonstrates that PersonSel had the most significant impact, and the result is consistent over two test sets. This finding suggests that teaching the LLMs to reason about the identity of the person in the summary could effectively enhance the model's ability to comprehend the conversation. 

Moreover, Correctness and SumGen also play a crucial role. On the one hand, for the model to decide whether a summary of the dialogue is correct, it should be capable of comprehending the information in the conversation accurately. On the other hand, as previously discussed, summarizing a dialogue involves understanding the dialogue, especially the salient information in it. Finally, even though SumSel shows some impact, it is relatively limited. We assume that this might be because SumSel has lengthy inputs, comprising a question, a dialogue and four summary, which could be challenging for the model to learn.


\section{Related Work}
\subsection{Dialogue Summarization and Question Answering}
Dialogue summarization is an important task in natural language generation~\cite{dialogueDataSurvey}, which involves summarizing the main content of a dialogue into a third-person summary. It can be used for various scenarios such as meetings~\cite{ami,zhong2021qmsum}, customer service~\cite{didi}, and daily chats~\cite{samsum}. Due to the dispersed information in multi-turn utterances, abstractive summarization is naturally suitable for modeling dialogue summarization~\cite{data}.
There are some works that have proposed dialogue-based question answering datasets, such as DREAM~\cite{dream} which is derived from listening test data, and some~\cite{Yang2019FriendsQAOQ,Ma2018ChallengingRC} are question-answering data based on dialogue data such as TV drama scripts and constructed question by manual and rule-based methods. 

\subsection{Faithfulness in Dialogue Summarization}

Recent research has focused on factual consistency in dialogue summarization. \citeauthor{tang2021confit} proposed CONFIT, which identified and annotated several error types. Meanwhile, \citeauthor{huang2022ed} used this data to evaluate the effectiveness of BARTScore through unlikelihood training on negative samples. \citeauthor{data} identified shortcomings in CONFIT's methodology, scope, and analysis, re-identifying six types of inconsistencies. They annotated and analyzed the faithfulness of four BART fine-tuned dialogue summarization models and found that 35\% of the generated summaries contained errors. 

\subsection{Investigating Large Language Models}
In recent years, large language models represented by GPT-3~\cite{gpt3} have demonstrated impressive language capabilities across various NLP tasks, such as sentence classification, question answering, machine translation, among others. Some studies have fine-tuned large models on task-specific data to enable them to interact with humans and respond to their instructions effectively; such models include FlanT5~\cite{flan} and ChatGPT. Alpaca~\cite{alpaca} and Vicuna~\cite{vicuna} collected interaction data from large model APIs to construct the instruction dataset, and they trained impressive instruction-following models. While these models are flourishing, some research has explored the task capabilities of large models. For example, \citet{chatgptEval} conducted a simple comparison of 20 datasets across 7 NLP tasks. However, only ChatGPT was included and only ROUGE~\cite{rouge} metrics were compared. In contrast, the aim of this study is to provide a deeper understanding of the dialogue comprehension capability.

\section{Conclusion}
In this paper, we conducted a detailed evaluation and analysis of the factual consistency in dialogue comprehension of existing LLMs, and attempted to improve upon them. Specifically, we annotated the consistency of summaries generated by the LLMs, and constructed factual questions from generated errors to evaluate the existing models. The evaluation results showed that there are still serious dialogue comprehension defects in current LLMs. Therefore, we made some initial attempt by using automatically constructed multi-task pseudo data to fine-tune the model. The experimental results showed that after fine-tuning, the model's dialogue comprehension capabilities were indeed enhanced, providing a possible solution for further research.

\section{Acknowledgments}
We would like to thank the anonymous reviewers for their insightful comments. Shujian Huang is the corresponding author. This work is supported by National Science Foundation of China (No. 62376116, 62176120), the Liaoning Provincial Research Foundation for Basic Research (No. 2022-KF-26-02), research project of Nanjing University-China Mobile Joint Institute.

\section*{Limitations}

Our data is manually annotated, which inevitably introduces certain biases. However, we have employed double annotation confirmation to minimize these biases as much as possible.
Our study primarily focuses on the dialogue understanding abilities of LLMs, rather than covering all aspects. This means that our conclusions may not fully represent the performance of LLMs in other tasks or domains. We need to be cautious about its misuse and also need more analysis of the abilities of LLMs in the future. Meanwhile, due to resource constraints, this study mainly evaluated relatively popular large models and was unable to cover all existing large models.
Finally, as a preliminary attempt to improve general dialogue understanding ability, the data construction strategy we used is relatively simple. Further in-depth research is needed to improve the dialogue understanding capabilities of LLMs in the future.

\section*{Ethics Statement}
The authors have no conflicts of interest. The datasets used come from publicly available sources and are compliant with their published license. And all data used and created in this paper does not include sensitive content such as personal information in real world. LLMs were only used to generate summaries and answer factual questions, in accordance with their intended use. 
\bibliography{custom}

\appendix
\clearpage
\section{Appendix}

\subsection{Filtering Rules for Summary Generation}\label{sec: filter rule}
When collecting summaries from LLMs, we noticed that some models generate results that do not contain exact information of the conversation, such as using the vague term "the group" to refer to people, and using words such as "the dialogue" to start a vague summary. These summaries cannot reflect the level of the model's understanding of the conversation, and are not suitable for the evaluation of factual consistency. Therefore, if the model produces a result that includes any of the phrases listed in Table~\ref{tab:rule used}, we reject the result and prompt the model to generate new one.

 \subsection{Annotation Details}\label{sec: Annotation Details}
 We recruited five annotators and explicitly informed them that the purpose of the annotated data is to evaluate the conversation comprehension ability of large language models. We fistly presented and explained each error types defined in \citealp{data} and its examples for each annotator. Then we instructed the annotators to annotate using the following guidelines: ``Read the dialogue carefully. Determine whether the summary contains factual inconsistency based on the dialogue. If so, identify the category of error and correct it. For each error, write three questions with as much variety as possible. Use the corrected summary to answer the questions as the correct option, and the erroneous summary to answer the questions as the distractor.'' During the entire annotation process, the payment for the annotators was sufficient. The use and public access of the data has been approved, and we will make the data publicly available after the end of the anonymous period to facilitate future research.

\begin{table}[hp]
\setlength{\tabcolsep}{1mm}
\centering
\footnotesize
\begin{tabular}{l}\toprule
\textbf{Phrases}\\\midrule\midrule
"In this dialogue", "In the dialogue",  "In this conversation",\\ "In the conversation","The conversation revolves",\\ "The dialogue", "The speaker","The group",\\"A group","The conversation"
\\\midrule 
\end{tabular}
\caption{\label{tab:rule used} When the generated summary contains one of these phrases, we will discard the results and ask the model to generate again. }
\end{table}

\subsection{Distribution of Error and Factual Question}\label{Distribution}
\begin{table}[htbp]
\setlength{\tabcolsep}{3mm}
\footnotesize
\centering
\begin{tabular}{lccccccl}\toprule
\textbf{Type} &  \textbf{E-Num}  & \textbf{Q-Num}& \textbf{Q-Num/E-Num}\\\midrule
\text{SubObjE} & 184 & 709 & 3.86 \\
\text{ProE}	&82	&282	&3.41\\
\text{HalE}	&99	&364	&3.68\\
\text{ParE}	&127&	441&	3.47\\
\text{NegE}	&59	&204	&3.45\\
\bottomrule
\end{tabular}
\caption{\label{ErrorDistribution} We report the number distribution of inconsistency error (E-Num) and our constructed factual question (Q-Num) over inconsistency types}
\end{table}

We performed a detailed statistical analysis to examine the distribution of errors and the associated factual questions across various categories. Our findings, as presented in Table~\ref{ErrorDistribution}, indicate a parallel trend between the distribution of questions and errors, with SubObjE emerging as the most common and NegE as the least. Upon assessing the mean number of questions elicited per error, it is noted that the disparity in the mean quantity of questions among different error categories is comparatively minimal. Notably, SubObjE consistently elicits the highest mean number of questions, whereas ProE is the lowest.

\subsection{General Ability is Preserved while Improving the Consistency}
In our study, we take the Alpaca-7B model as an example to investigate the impact of our fine-tuning strategy on on the general capability, while enhancing the consistency of dialogue understanding. To this end, we evaluated the model on the MTBench benchmark~\cite{mtbench} in Single-Score Mode and Pairwise Mode. 
In Single-Score Mode, the generated answers will be rated by GPT-4 on a scale from 1 to 10. In Pairwise Mode, GPT-4 will be presented with two answers from different models at the same time and required to select the superior one.
The experimental results indicate that the single score is increased from 4.86 to 5.02 after fine-tuning. Meanwhile, the pairwise comparisons show that the fine-tuned model produced better answers in 36.3\% of the cases, while only 33.7\% for the original model (with a 30\% tie rate). These findings suggest that our method not only effectively enhances the model's consistency in dialogue understanding but also successfully retains its general capability, demonstrating the efficacy and versatility of our approach.

\subsection{Instruction for Each Task}\label{sec:instruction}
Table~\ref{tab:instruction used} shows a subset of instruction templates that were used by us for training and evaluation purposes.
It can be observed that the training data instructions and the test instructions used have significant differences for different tasks, in both content and form. While the Correctness instruction requires answering yes or no questions, PersonSel and SumSel are four-option multiple-choice questions, and SumGen requires generating a summary. This design aims to enable the model to learn core conversational understanding abilities and avoid fitting shallow task formats only.

\begin{table*}[hp]
\setlength{\tabcolsep}{1mm}
\centering
\footnotesize
\begin{tabular}{r|p{12cm}}\toprule
\textbf{Task} & \textbf{Instruction} \\\midrule\midrule
\textbf{\SUM:}  & Please summarize the following conversation as brief as possible.\\ \midrule 
\textbf{\QA:}  & Here is a dialogue, a question, and two options below. Please read the dialogue carefully and\\
\textbf{DREAM:} &  select the best option to answer the question. Your answer should be either A or B.\\\midrule 
\textbf{QE:}  & 1. Please confirm whether the summary accurately reflects the participants involved in the event based on the conversation.\\
 & 2. Reviewing the summary, did anyone participate in the event is inconsistent with the dialogue? Yes or no answer please.\\
 & 3. According to the dialogue, are any of the person in the summary incorrect or missing? Please respond with yes or no. \\
 & \quad \dots \\\midrule 
\textbf{PersonSel}  & 1. Based on the conversation, select the most suitable option to fill in the [MASK] blank.  \\
& 2. After considering the discussion, which of the four choices would be the best to complete the [MASK] position?  \\
& 3. What is the most fitting option to complete the [MASK] blank based on the conversation?  \\
 & \quad \dots \\\midrule 
\textbf{SumSel} & 1. There is a dialogue below and four candidate summaries, of which three contain errors. Please select the correct one.\\
 & 2. Identify the correct summary of the dialogue from the provided options.\\
  & 3. From the given choices, choose the summary that correctly represents the dialogue without any errors.\\
 & \quad \dots \\\midrule 
\textbf{SumGen} & 1. Write a brief summary of this conversation\\
 & 2. Based on the dialogue, Summarize the main points\\
  & 3. Please provide a concise summary of the conversation\\
 & \quad \dots \\\bottomrule 
\end{tabular}
\caption{\label{tab:instruction used} Instructions used for generating the summarization and perform question-answering. We used the same instructions for tests in \QA and DREAM. For each training task, we randomly selected three instructions from a total of ten to present here. }
\end{table*}

\end{document}